\documentclass[acmtog,nonacm]{acmart}
\usepackage{booktabs} 
\usepackage{multirow}
\usepackage{tabularx}
\usepackage{wrapfig}
\usepackage{mathtools}
\usepackage{lipsum}
\usepackage{footmisc}
\usepackage{bbding}

\usepackage[ruled]{algorithm2e}

\SetAlFnt{\small}
\SetAlCapFnt{\small}
\SetAlCapNameFnt{\small}
\SetAlCapHSkip{0pt}

\begin{document}
\def\negativevspace{}

\newcommand{\para}[1]{\vspace{.05in}\noindent\textbf{#1}}
\def\ie{\emph{i.e.}}
\def\eg{\emph{e.g.}}
\def\etal{{\em et al.}}
\def\etc{{\em etc. }}
\newcolumntype{C}[1]{>{\centering\arraybackslash}p{#1}}

\title{CNS-Edit: 3D Shape Editing via Coupled Neural Shape Optimization}

\author{Jingyu Hu}
\affiliation{%
	\institution{The Chinese University of Hong Kong} \country{HK SAR, China}}
\author{Ka-Hei Hui}
\affiliation{%
	\institution{The Chinese University of Hong Kong} \country{HK SAR, China}}
\author{Zhengzhe Liu}
\affiliation{%
	\institution{The Chinese University of Hong Kong} \country{HK SAR, China}}
 \author{Hao (Richard) Zhang}
\affiliation{%
	\institution{Simon Fraser University} \country{Canada}}
\author{Chi-Wing Fu}
\affiliation{%
	\institution{The Chinese University of Hong Kong} \country{HK SAR, China}}
\renewcommand\shortauthors{Hu el al.}

\begin{abstract}
%
This paper introduces a new approach based on a coupled representation and a neural volume optimization to implicitly perform 3D shape editing in latent space.
This work has three innovations.
First, we design the coupled neural shape (CNS) representation for supporting 3D shape editing.
This representation includes a latent code, which captures high-level global semantics of the shape, and a 3D neural feature volume, which provides a spatial context to associate with the local shape changes given by the editing. 
Second, we formulate the coupled neural shape optimization procedure to co-optimize the two coupled components in the representation subject to the editing operation. 
Last, we offer various 3D shape editing operators,~\ie, copy, resize, delete, and drag, and derive each into an objective for guiding the CNS optimization, such that we can iteratively co-optimize the latent code and neural feature volume to match the editing target.
With our approach, we can achieve a rich variety of editing results that are not only aware of the shape semantics but are also not easy to achieve by existing approaches.
Both quantitative and qualitative evaluations demonstrate the strong capabilities of our approach over the state-of-the-art solutions.

\end{abstract}

\maketitle

\begin{figure}[t]
	\centering
	\includegraphics[width=0.99\columnwidth]{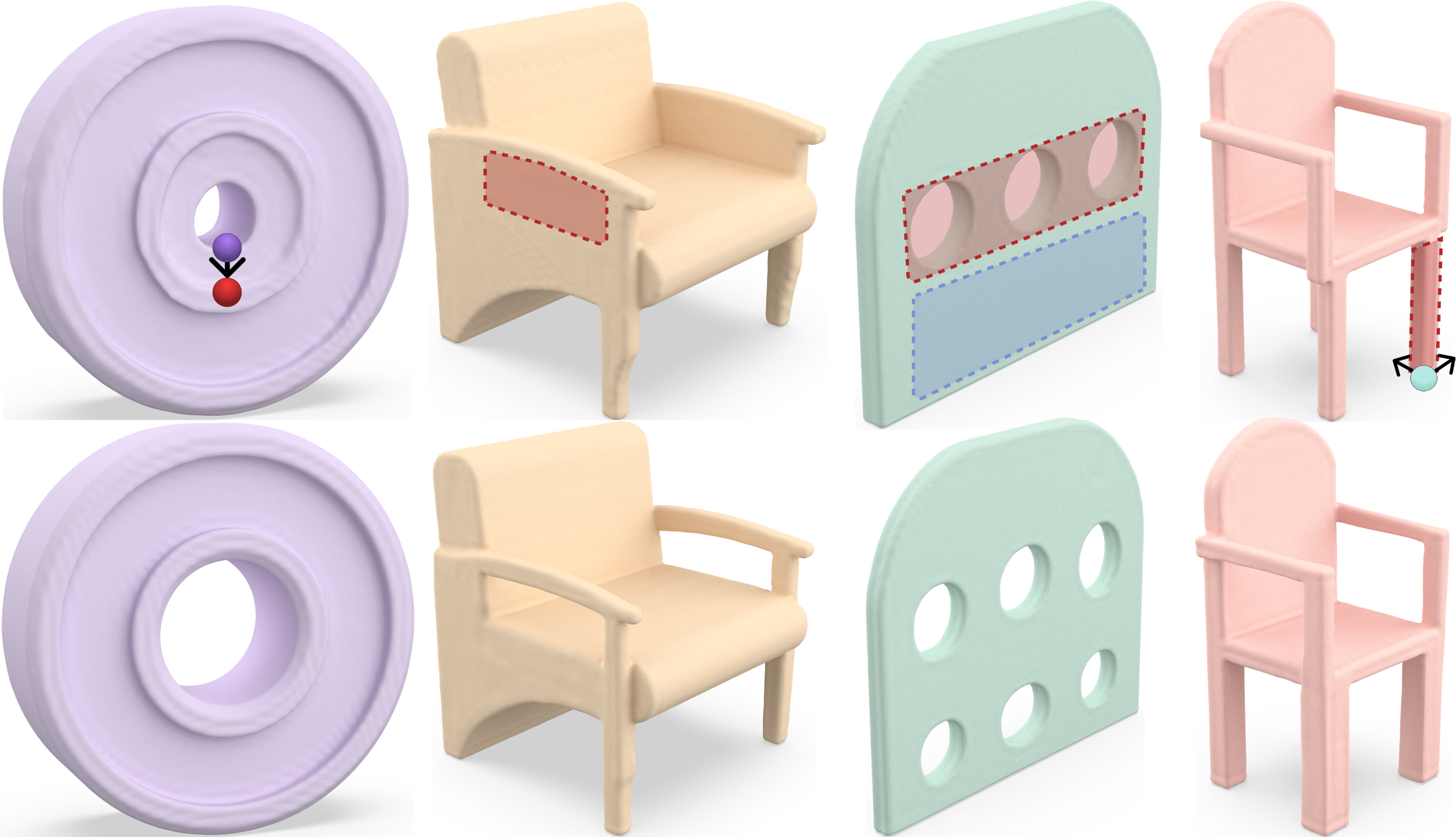}
        \vspace{-2mm}
	\caption{
        We propose a novel coupled neural shape representation, equipped with a 
        family of user-friendly shape editing operators: (i) drag (first column), (ii) delete (second column), (iii) copy (third column), and (iv) resize (fourth column). The top row shows the input shapes and operators, whereas the bottom row shows the edited results.
     }
	\label{fig:teaser}
        \vspace{-5mm}
\end{figure}
\section{Introduction}
\label{sec:intro}

As one of the most classical and foundational problems in computer graphics, 3D shape editing has been receiving renewed interests in the realm of 
representational learning and neural geometry processing lately. Often, the key lies in developing a suitable neural shape representation, which
enables: (i) a variety of operators to be developed for intuitive and fine-grained editing; (ii) an understanding of shape semantics to produce 
natural and accurate responses to the editing operation; and
(iii) preservation of unedited 
shape regions and features, while ensuring high-quality edit outcomes.

Classical shape editing and manipulation typically works with some form of editing {\em handles\/} or {\em proxies\/}~\cite{yuan2021revisit},~\eg, 
points, curves, sketches, skeletons, or cages. Such proxies often provide a reduced shape representation, or an abstraction, which are easier
to manipulate. With many deep neural representations developed for 3D shapes, a common approach to shape editing is to perform the operations
in a {\em latent space\/}, which acts like a proxy, and is often tied to a generative model such as autoencoders, adversarial networks (GANs)~\cite{goodfellow2014generative}, 
or diffusion~\cite{ho2020denoising}. Latent-space editing has seen much success for images~\cite{pan2023drag,shi2023dragdiffusion,mou2023dragondiffusion} 
and videos~\cite{deng2023dragvideo}, where the editing mainly involves object deformation or scene layout alterations through dragging.
As Figure~\ref{fig:teaser} shows, typical 3D shape edit operations are of a rather different nature.
Recent methods for latent-space 3D shape manipulations~\cite{hertz2020pointgmm,hao2020dualsdf,hui2022template,koo2023salad} seek to couple implicit functions with 
geometric primitives or connect a shape latent space with CLIP space~\cite{hu2023clipxplore}. 
However, due to the coarseness of the primitives or text prompts, they often struggle 
with edit quality, especially when fine-grained editings are performed.

In this paper, we introduce a novel {\em coupled neural shape\/} (CNS) representation for 3D shapes, which is semantic-aware and facilitates a variety of intuitive and fine-grained 
edit operations. Our CNS representation includes a latent code, which captures high-level global semantics (e.g., symmetry) of a given shape, and a 3D 
neural feature volume, which provides a spatial context to associate with the local shape changes given by an edit. The edits are {\em implicitly\/} performed in the latent
space through a 
{\em coupled neural shape optimization\/}.
Specifically, we 
co-optimize the two coupled components in the 
CNS representation subject to an editing operation being performed. The editing operators we currently offer include copy, resize, delete, and drag, and each is derived into an 
objective for guiding the CNS optimization, such that we can iteratively co-optimize the latent code and neural feature volume to match the editing operation.

With our approach, which is coined CNS-Edit, we can achieve a rich variety of editing results that are not only aware of the shape semantics but are also not easy to achieve by 
any existing approaches. For example, as demonstrated in Figure~\ref{fig:teaser}, both the delete and copy operators can introduce topology changes to a 3D shape, which are
seamlessly accomplished by our method. Both quantitative and qualitative evaluation results demonstrate the strong capabilities of CNS-Edit over the state-of-the-art solutions.

\vspace{-1mm}
\section{Related Work}
\label{sec:rw}

Shape editing has been a long-standing challenge in graphics.
To maintain geometric fidelity in shape editing, researchers utilize a variety of techniques,~\eg, including but not limited to ARAP deformation~\cite{sorkine2007rigid}, cage-based deformation~\cite{joshi2007harmonic, ju2005mean, lipman2008green}, differential coordinates~\cite{lipman2004differential} and laplacian operator~\cite{sorkine2004laplacian}.
These early shape editing techniques all operate at the level of mesh vertices and lack semantic understanding or control. 

Then came the era of {\em structure-aware\/} shape manipulation~\cite{mitra_star13}, which generally follows an analyze-and-edit paradigm to first extract the structures of
an edited shape as reflected by feature curves~\cite{gal2009iWire}, part bounding boxes~\cite{zheng2011controller}, and symmetries~\cite{wang2011symh}, and
then perform edits to preserve these structures. Yet, before the advent of deep learning, the capability of the structure discovery schemes were limited and brittle. Also,
when working with segmented 3D shapes with separated parts, maintaining proper part connections has been an often overlooked challenge~\cite{yin20203dv}.
In this section, we mainly focus on learning-based approaches for shape generation, manipulation, and editing, which are most relevant to our work.

\vspace{-1mm}
\paragraph{Shape Editing with Learned Models.}
One line of approaches~\cite{hertz2020pointgmm, hao2020dualsdf, hui2022template, koo2023salad} has focused on coupling implicit functions with basic geometric primitives,
allowing for the editing of the former by modifying parameters of the latter.
Such approaches for
shape generation
enhances the flexibility, but typically at the expense of generative quality. 
Other methods~\cite{wang20193dn, yifan2020neural, liu2021deepmetahandles, jiang2020shapeflow} developed deformation-based shape editing. 
Yet, a notable limitation of these methods is the lack of capability to edit the topology of shapes.
Recent works also try to involve additional modalities such as texts and sketches for more %
shape manipulation. 
Text-guided methods~\cite{liu2023exim, fu2022shapecrafter, liu2022towards, achlioptas2022changeIt3D, huang2022ladis} excel in high-level semantic modifications but often fall short in providing precise spatial control during editing. While sketch-guided approaches~\cite{gao2022sketchsampler, zhang2021sketch2model, guillard2021sketch2mesh, zheng2023lasdiffusion, hu2023clipxplore} allow users to alter a shape by modifying its sketches, they often do not accommodate non-expert (hence low-quality) or out-of-distribution inputs.
Of note is CLIPXplore~\cite{hu2023clipxplore}, which also designs a coupled representation and co-optimization. By connecting a CLIP space with a 3D shape space, their representation targets shape {\em exploration\/}, not fine-grained editing.
Our CNS-Edit introduces a new coupled neural shape representation for 3D shape editing with a set of operators to manipulate this representation. These operators are designed to facilitate efficient and intuitive 3D shape editing,
offering high-fidelity results, fine-grained controllability, and topology modification.

\vspace{-1mm}
\paragraph{3D Shape Generation via Different Representations}
Many methods have explored 3D generation using classical shape representations or emerging neural representations. These include voxels~\cite{wu2016learning, smith2017improved}, point clouds~\cite{zhou20213d,luo2021diffusion,zeng2022lion, nichol2022point, li2021spgan, gal2020mrgan, hui2020progressive}, meshes~\cite{Liu2023MeshDiffusion, siddiqui2023meshgpt, chen2020bsp},
implicit functions~\cite{hui2022neural,gao2022get3d,zhang20233dshape2vecset,mescheder2019occupancy,chen2019learning,hao2020dualsdf,ibing20213d,chou2023diffusion}, and neural radiance fields (NeRF)~\cite{mildenhall2021nerf,lin2023magic3d,poole2022dreamfusion}.
While these advanced works excel in high-fidelity shape generation, they were not designed for fine-grained editing tasks, as they 
typically
lack sufficient controllability. %

\vspace{-1mm}
\paragraph{Image Editing via Generative Models}
Generative adversarial networks (GANs)~\cite{goodfellow2014generative} have been foundational in the development of many subsequent methods in the field of image manipulation and editing~\cite{abdal2021styleflow, goetschalckx2019ganalyze, shen2020interpreting, shen2021closed, voynov2020unsupervised, cherepkov2021navigating}.
Despite the advancements in GANs, a significant challenge arises in accurately inverting real images back into GAN latent codes. This challenge is attributed to the limitations in their generative capabilities~\cite{abdal2019image2stylegan}. 

Consequently, these constraints significantly hinder the generalization ability of GANs in various real-world image editing tasks.
Recently, with the advent of large-scale text-to-image diffusion-based models like~\cite{rombach2022high}, many diffusion-based methods~\cite{bar2022text2live, brooks2023instructpix2pix, hertz2022prompt, kawar2023imagic, parmar2023zero} are proposed for text-based image editing.
However, it is still an open problem to edit images using the text, since texts typically lack precise, pixel-level spatial controllability.
Most recently, Pan et al.~\shortcite{pan2023drag} proposed ``Drag Your GAN'' to further enhance image editing controllability.
It enables pixel-precise editing by allowing users to interactively drag any image points to target locations, yielding impressive results.
Following~\cite{pan2023drag}, subsequent studies~\cite{shi2023dragdiffusion, mou2023dragondiffusion} have expanded the manipulation framework into the stable-diffusion~\cite{rombach2022high}. This extension significantly enhances the quality of image manipulation. 
Yet, there still lacks an effective framework in the 3D domain for shape editing that simultaneously maintains semantic and precise spatial controls.

\begin{figure*}[!t]
	\centering
	\includegraphics[width=2.1\columnwidth]{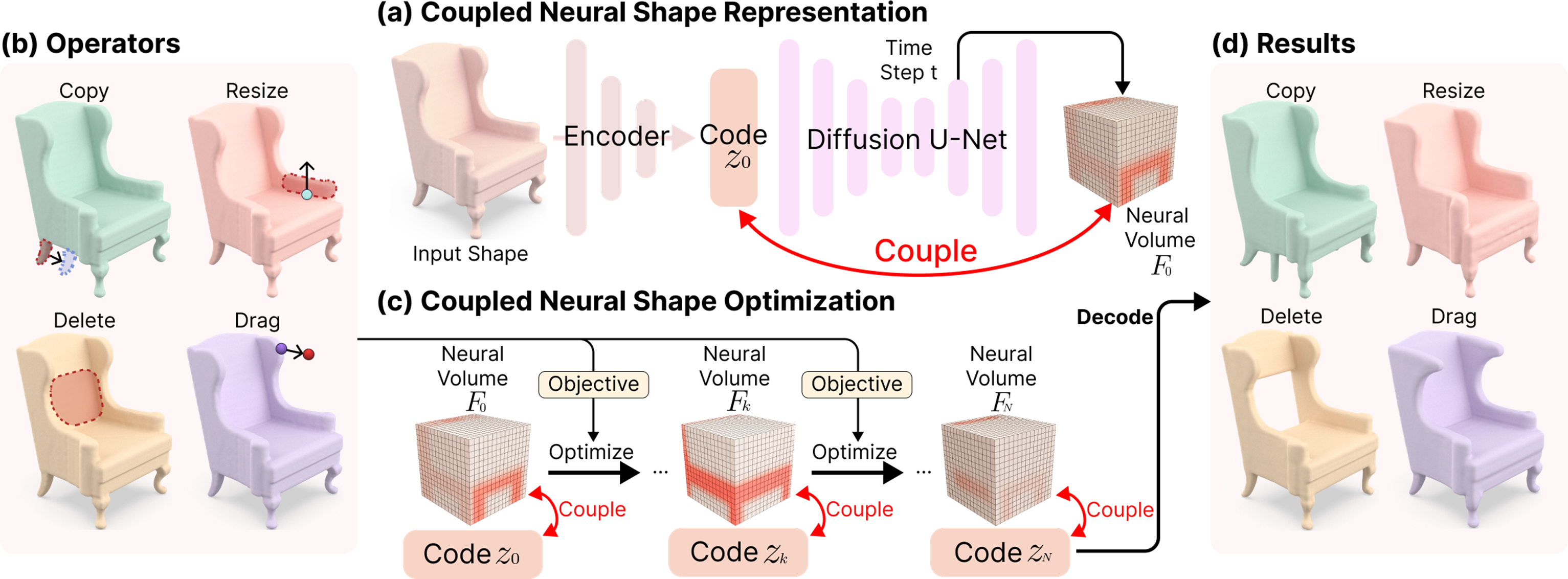}
        \vspace{-6mm}
	\caption{
     Overview of our framework. (a) We propose a new coupled neural shape (CNS) representation, consisting of latent code $z$ and neural feature volume $F$.
     From a given shape, we first adopt an encoder network to derive its global latent code $z$.
     Then code $z$ is fed into the Diffusion U-Net to extract intermediate features, from which we obtain the neural feature volume $F$. Notice that code $z$ and neural volume $F$ are closely coupled.
     Next, we provide (b) a family of operators,~\ie, copy, resize, delete, and drag, for shape editing, and
     (c) transform the operator into an objective for guiding the iterative co-optimization of $z$ and $F$.
     After $N$ iterations of co-optimization, (d) we can obtain the updated latent code $z_N$ and decode it to produce the edited shape.}
     \vspace{-0.95mm}
	\label{fig:overview}
\end{figure*}

\vspace{-1mm}
\section{Overview}
\label{sec:overview}

Given a 3D shape, our goal is to modify it according to an editing operation.
Specifically, we introduce the Coupled Neural Shape (CNS) representation, allowing us to take advantage of a pre-trained latent shape space, such that we can 
modify the shape implicitly through the CNS representation in the latent space to match the editing operation. 
Our framework is illustrated in Figure~\ref{fig:overview}.

There are three key innovations in this work:
\begin{itemize}
\item
First, we design the Coupled Neural Shape (CNS) representation, which consists of global latent code $z$ and neural feature volume $F$, in which code $z$ provides the global 
shape semantics
whereas the neural volume $F$ provides the spatial context for associating with the editing operation.
The two components are closely coupled from one another and complement each other to aid the shape editing.
Figure~\ref{fig:overview} (a) overviews the general procedure of constructing them from a given shape; see Section~\ref{subsec:feature_extraction} for the details. 
\item
Second, we devise the coupled neural shape optimization procedure 
(Section~\ref{subsec:neural_volume_op}) to modify the CNS representation for the shape editing operation.
Specifically, we first translate the editing operation into an optimization objective defined on the neural volume $F$; then, as Figure~\ref{fig:overview} (c) shows, we can iteratively co-optimize code $z$ and neural volume $F$ to produce a refined code $z_N$.
Importantly, in each iteration, we search for code $z_k$
using gradient descent in the shape space, such that its associated neural volume $F_k$ follows the editing objective.
In this way, we can ensure structural integrity of the edited shapes while considering the shape semantics.
\item
Last, we introduce a family of editing operators,~\ie, copy, resize, delete, and drag (Section~\ref{subsec:volume_operator}); see Figure~\ref{fig:overview} (b) for the illustrations, and derive procedures to translate each of them to an operator-specific objective (Section~\ref{subsec:op_objective}).
Using such objective, we can appropriately guide the co-optimization on the CNS representation to produce the final code $z_N$ and decode $z_N$ to produce the edited shape; see Figure~\ref{fig:overview} (d).
\end{itemize}

\vspace*{-1mm}
\section{Method}
\label{sec:architecture}

\subsection{Coupled Neural Shape Representation}
\label{subsec:feature_extraction}
To begin, we first introduce the Coupled Neural Shape (CNS) representation, designed for enabling 3D shape editing.
This representation is derived from the input shape, consisting of a pair of \textit{coupled} neural tensors:
(i) a {\em global latent code\/} $z$, which captures the overall shape semantics,
and
(ii) a {\em 3D neural feature volume\/} $F$, which captures the spatial context, such that we can associate spatial locations between the neural volume and input shape.
Before we present the procedure to obtain the CNS representation, we first introduce the following two building blocks for creating the representation:

\para{(i) Input Shape Representation.} \
Each input shape is first encoded into a compact wavelet coefficient volume, following~\cite{hui2022neural}. Specifically, we sample each shape into a high-resolution truncated signed distance field volume and convert the distance field into a wavelet coefficient volume.
Further, it is worth to note that this wavelet coefficient volume maintains a spatial association with the original shape. Thanks to the {\em local support\/} property of wavelet transform. A local change in the wavelet volume affects only an associated local region of the shape, and vice versa.

\para{(ii) Semantic Shape Latent Space.} \
Another building block is a pre-trained shape latent space, to which we can encode the input shape. In our setting, we employ the pre-trained diffusion-based autoencoder from~\cite{hu2023neural}, adopting its encoder to produce the global latent code
and its decoder to reconstruct the original shape volume from the latent
code.
The decoder is formulated using the denoising diffusion probabilistic model~\cite{ho2020denoising}.  In short, after 
sampling a 3D noise volume, we pass it together with the global latent code to a U-Net structure to denoise the volume.  By iteratively denoising the volume through the U-Net for $T$ iterations, we can obtain the decoded shape volume.

\vspace*{2.6mm}
Using the two building blocks, we can prepare the two neural tensors in our CNS representation using the following two steps:

\para{Step (i): Extract the Global Latent Code $z$.} \
We first pass the input shape, represented as a wavelet coefficient volume, to the pre-trained encoder to obtain an initial latent code $z'$.
We then fine-tune code $z'$, following the optimization process in~\cite{hu2023neural}, to produce a more faithful latent code $z$. Importantly, this refined
code $z$ provides a semantically more meaningful latent space.
So, by leveraging this space to form the global latent code component in our CNS representation, we can effectively pinpoint new latent code and produce new shape for the editing operation.

\para{Step (ii) Extract the Neural Feature Volume $F$.}
The global latent code $z$ alone is insufficient for supporting shape editing.
It lacks spatial context to associate with specific spatial changes in the shape that the editing operation targets.
Hence, we propose a coupled representation by additionally constructing a 3D neural feature volume, whose spatial context is coupled with $z$.
To do so, we intentionally perform $t$ iterations with the U-Net in denoising the noise volume, where $t < T$. 
We then feed the partially-denoised volume and the global latent code $z$ again into the U-Net and extract intermediate features from the U-Net.
Specifically, we take the feature volume in the fourth last layer of the U-Net as the 3D neural feature volume $F$ in our CNS representation.
This choice is made, as the feature volumes in deeper layers do not have sufficient spatial context, whereas those from shallower layers lack shape semantics.
Importantly, this volume is further processed only through convolution layers with a limited spatial receptive field before producing the final output.
Therefore, local modifications in this volume typically correspond to targeted changes within the desired regions in the shape.

\para{Note: $F$ and $z$ are Coupled.} \
It is important to highlight that the two neural tensors in our CNS representation are closely coupled.
We can induce changes on one component, based on modifications on the other.
On the one hand, inducing changes on the neural volume $F$ from a modified $z$ is straightforward, by following 
the procedure in Step (ii) above.
On the other hand, given a modified neural volume $F$, since the extraction of $F$ is differentiable,
we measure the changes between the original volume and the modified one using a loss.
So, we can backpropagate the changes to the global code $z$ and use gradient descent to obtain the updated code $z_N$ that better matches the modified $F$, as Figure~\ref{fig:overview} (c) illustrates. With these two coupling relations, the shape editing task can be posed as how to derive the objective based on the editing operation and then co-optimize (optimize together) the neural volume $F$ and code $z$ accordingly.

\begin{figure}[!t]
	\centering
	\includegraphics[width=0.99\columnwidth]{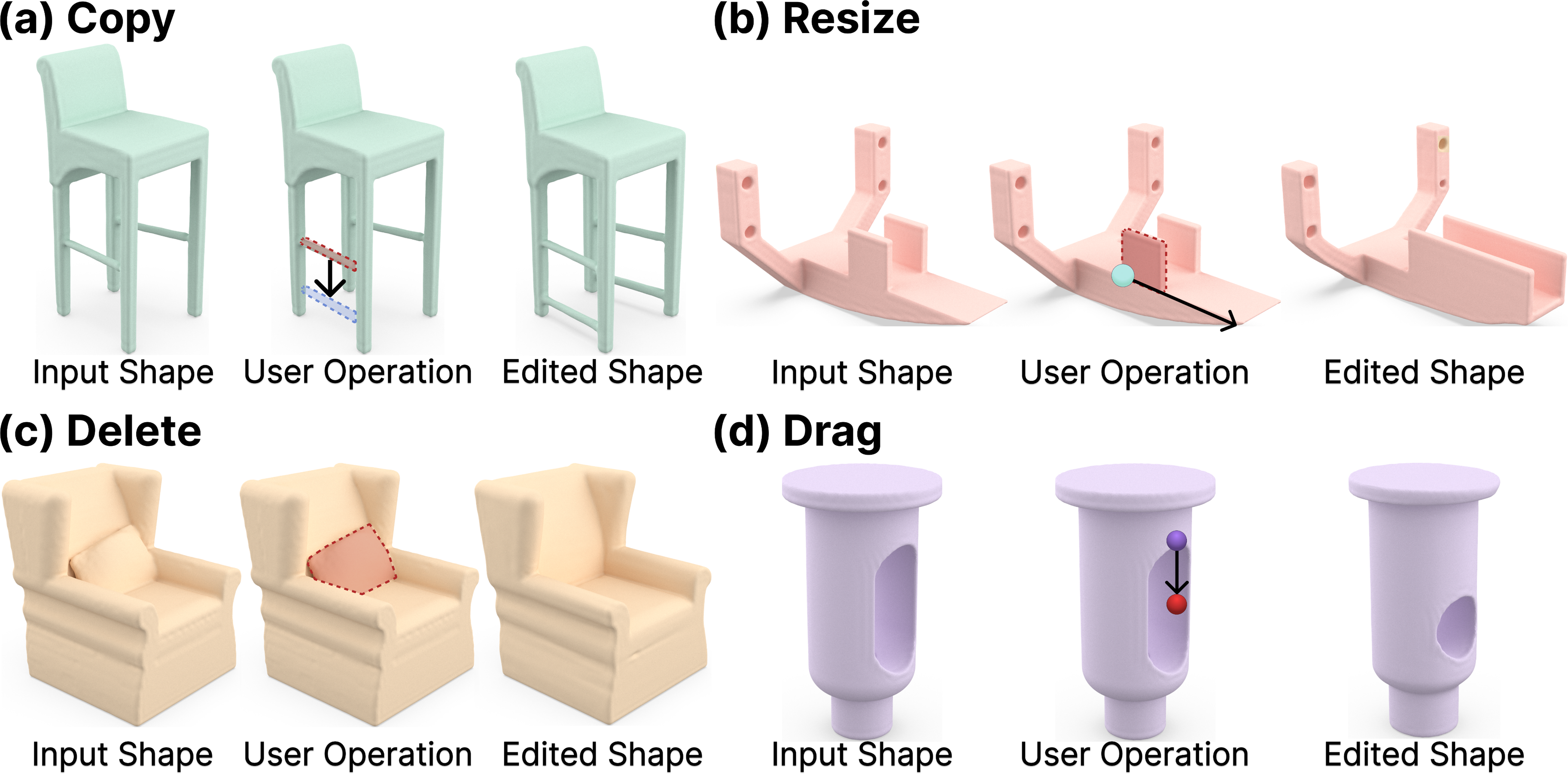}
        \vspace{-2mm}
	\caption{
 Shape editing operators: copy, resize, delete, and drag.
 Note the fidelity of the edited shapes produced by our method.
 }
	\label{fig:demo_op}
\end{figure}

\subsection{Shape Editing Operators}
\label{subsec:volume_operator}
In this work, we propose the following shape editing operations:
\begin{itemize}
\item[(i)] \para{Copy Operator.}\ 
This operator enables one to copy a portion of the input shape and paste it to some other location in the shape.
As illustrated in Figure~\ref{fig:demo_op} (a), one can mark a source region (in red) and specify a displacement (indicated by the arrow) from the source region to the target region (in blue).
Then, the operator can copy the local geometry from the source to the target region.
It is important to note that the pasted geometry will be seamlessly integrated with the original target region in the shape; see,~\eg, the cloned stretcher that connects with the chair legs in Figure~\ref{fig:demo_op} (a).
\item[(ii)] \para{Resize Operator.}\ 
This operator enables one to scale a selected region in a shape about a chosen anchor point along desired direction(s).
As Figure~\ref{fig:demo_op} (b) shows, the chosen region is marked in red; the desired direction is marked by the black arrow; and the anchor point is marked by the blue dot.
Using resize, we can scale a local part along a specific direction; see,~\eg, the rectangular panel in the CAD model shown in Figure~\ref{fig:demo_op} (b).
Note particularly in this example that we mark only the left panel.
The right panel in the CAD model can be automatically resized, following the left panel.
\item[(iii)] \para{Delete Operator.}\ 
This operator enables one to remove a selected part in the input shape.
For example, as Figure~\ref{fig:demo_op} (c) shows, one can mark the cushion on the couch (in red) and proceed to remove it. It is noteworthy that once the cushion is removed, the couch's revealed back is semantically made coherent with the surrounding regions in the shape.
\item[(iv)] \para{Drag Operator.}\ 
Motivated by~\cite{pan2023drag}, this operator enables one to drag the local geometry around a point toward a target location, such that the geometries along the path can be adapted to the changes induced by the drag.
As Figure~\ref{fig:demo_op} (d) shows, by just marking and dragging a source point (in purple) towards a target point (in red), we can reduce the size of the hole.
Note that this operation is performed just by marking a point and dragging it.
\end{itemize}
\vspace{-1mm}

\begin{figure}[!t]
	\centering
	\includegraphics[width=0.92\columnwidth]{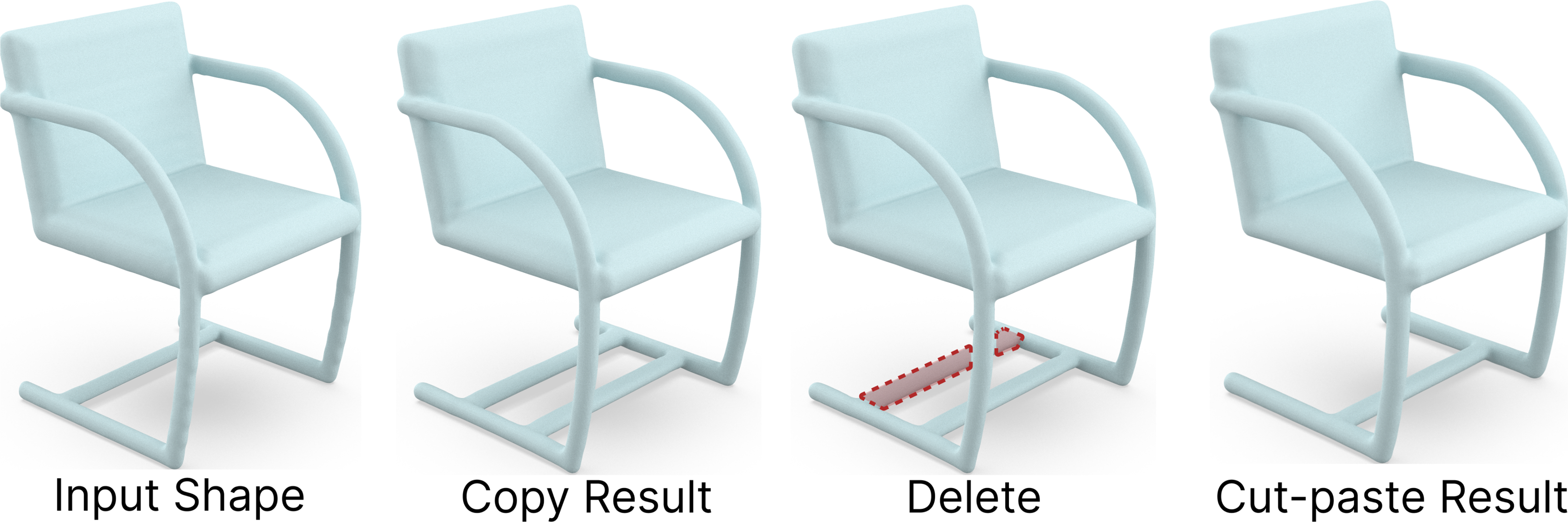}
        \vspace{-2mm}
	\caption{The cut-paste operator combines the copy and delete operators.}
	\label{fig:demo_cut}
        \vspace*{-1.5mm}
\end{figure}

\vspace*{2mm}
Though some of the above operators are partially achievable by traditional methods,~\eg, resize can be done in~\cite{joshi2007harmonic, ju2005mean, lipman2008green}, 
our method considers the editing operation more semantically.
For example, when resizing a part of the shape, associated symmetric part(s) can be automatically updated; see,~\eg, Figure~\ref{fig:demo_op} (b).
This result demonstrates the shape semantics that our method has considered in the editing process,~\eg, symmetry.
Results for various operators are provided in Figures~\ref{fig:teaser},~\ref{fig:gallery_resize},~\ref{fig:gallery_copy},~\ref{fig:gallery_delete}, and ~\ref{fig:gallery_drag}.
Also, the operators introduced above are atomic, allowing new operators to be formed by combining the existing ones.
For instance, combining ``copy'' and ``delete'' results in the ``cut-paste'' operator.
As Figure~\ref{fig:demo_cut} shows, we first replicate (copy operator) the horizontal stretcher of a shape using the copy operator, then we can remove (delete operator) the original stretcher in the shape.

\subsection{Coupled Neural Shape Optimization}
\label{subsec:neural_volume_op}

Next, we introduce the overall procedure, the coupled neural shape optimization,
to modify the CNS representation for a given editing operation.
This procedure has two major steps.
First, we derive an operator-specific objective $\mathcal{L}_{\text{op}}$ from the editing operation, where $\mathcal{L}_{\text{op}}$ specifies how the neural feature volume $F$ of the input shape should be updated.
Second, using objective $\mathcal{L}_{\text{op}}$ as the guidance, we co-optimize the two components in the CNS representation and further decode the optimized CNS to produce the edited shape.

\para{(i) Derive objective $\mathcal{L}_{\text{op}}$ from editing operation.}
Objective $\mathcal{L}_{\text{op}}$ contains two parts.
The first part is a list of spatial coordinates in the neural feature volume $F$, specifying the target region in $F$ that the editing operation aims to induce changes:
coordinate list $\Gamma = \{(x_i, y_i, z_i) | \forall i \in [1, \cdots,M]\}$, where M is the length of the list.
The second part is a list of target feature values associated with the spatial coordinates in the first list:
value list $V = \{f_i | f_i \in \mathbb{R}^C, \forall i \in [1, \cdots, M] \}$, where $C$ is the channel size.
Using the two lists, we can then formulate the operator-specific objective $\mathcal{L}_{\text{op}}$ as
\begin{equation}
\label{eq:op_loss}
\mathcal{L}_{\text{op}} = | F_k[\Gamma] - sg(V) |_{1},
\end{equation}
where $F[\cdot]$ represents the slicing operation on the neural feature volume $F$; $|\cdot|_1$ is the standard $L_1$ loss; and $k$ is the iteration round.

The above formulation is general for all the operators presented in Section~\ref{subsec:volume_operator}.
Section~\ref{subsec:op_objective} will provide details on how to derive objective $\mathcal{L}_{\text{op}}$,~\ie, $\Gamma$ and $V$, for each operator.
Also, if a location in $\Gamma$ is not an integer coordinate in $F_k$, the output of the slicing operation is obtained by a trilinear interpolation on the neighborhood values.
Besides, we employ the stop-gradient operator $sg$ on the target values in $V$ to 
encourage the values in $F_k$ to approach the target values, rather than the other way around, in the optimization.

\para{(ii) Coupled Co-optimization.}
Once $\mathcal{L}_{\text{op}}$ is defined, we co-optimize latent code $z$ and neural volume $F$ in the CNS representation for $N$ iterations accordingly.
At the beginning of the co-optimization, we first compute the starting neural volume $F_0$ based on the global latent code $z_0$ of the input shape, following the procedure described in Section~\ref{subsec:feature_extraction}.
We then evaluate $\mathcal{L}_{\text{op}}$ on the produced neural volume $F_0$ to measure the difference between the values within the target region $\Gamma$ of the neural volume $F_0$ and the desired target values $V$.
By evaluating the gradient of $\mathcal{L}_{\text{op}}$  with respective to $z_0$, we can produce an updated $z_1$ by taking one gradient step with a learning rate of $\alpha$. 
By repeating the above procedure on the newly produced latent 
code, we can obtain a sequence of CNS representation,~\ie, $\{(z_0, F_0), (z_1, F_1), \cdots, (z_N, F_N)\}$, as illustrated in Figure~\ref{fig:overview} (c).

Overall, 
the updated CNS representation at later optimization steps should better follow the editing operation.
Even if we define a maximum number of optimization iterations, the optimization may stop early, if a termination condition is reached; see Section~\ref{subsec:imp_detail} for the details on the termination condition defined for individual operators.
Last, the optimized CNS representation can then be decoded to produce the edited shape by running the decoder with the remaining time steps, as guided by the optimized latent code $z_N$.

\subsection{Deriving Objective for Each Operator}
\label{subsec:op_objective}
This section presents the operator-specific objectives, $\mathcal{L}_{\text{copy}}$, $\mathcal{L}_{\text{resize}}$, $\mathcal{L}_{\text{delete}}$, and $\mathcal{L}_{\text{drag}}$.
Primarily, we define appropriate target regions $\Gamma$ and target values $V$ for each operator for constructing Eq.~\eqref{eq:op_loss}:

\begin{itemize}

\item[(i)] \para{$\mathcal{L}_{\text{copy}}$.}\
Given the selected region to be copied, we first determine the coordinates in the neural volume covered by the selection to construct the coordinate list $\Gamma_{\text{copy}}$. We then determine the pasting region $\Gamma_{\text{paste}}$ by shifting each coordinate in $\Gamma_{\text{copy}}$ by the user-provided displacement vector.
Our goal here is to encourage the feature values in $\Gamma_{\text{paste}}$ to closely match the associated values in $\Gamma_{\text{copy}}$.
So, we set $\Gamma=\Gamma_{\text{paste}}$ and $V = F_0[\Gamma_{\text{copy}}]$ to form the objective $\mathcal{L}_{\text{copy}}$.

\item[(iii)] \para{$\mathcal{L}_{\text{resize}}$.}\
Given the selected region to be resized, we denote the associated selected coordinates set as $\Gamma_{\text{resize}}$. Then, we find bounding box $\mathcal{B}$
that encompasses the entire $\Gamma_{\text{resize}}$ and resize $\mathcal{B}$ into a new bounding box $\mathcal{B}'$ based on the specified anchor point and resize direction.
So, the target region $\Gamma$ is defined as the set of coordinates inside bounding box $\mathcal{B}'$. 
As for the target feature values $V$, they are located in $\mathcal{B}'$ and should proportionally come from 
those in $\mathcal{B}$.
So, we use trilinear interpolation to sample feature values in $\mathcal{B}$ of $F_0$ to obtain the target feature value at each coordinate in $\Gamma$.

\item[(iii)] \para{$\mathcal{L}_{\text{delete}}$.}\
Given the selected region to be removed, we denote the coordinates within this region as $\Gamma_{\text{delete}}$.
Our idea is to adjust the feature values in $\Gamma_{\text{delete}}$ to match those in the empty region of neural volume $F_0$. 
Hence, we look for an empty region, denoted as $\Gamma_{\text{empty}}$, in the given shape, and extract the local feature values. 
Then, we set $\Gamma=\Gamma_{\text{delete}}$ and $V=F_0[\Gamma_{\text{empty}}]$ to form the objective $\mathcal{L}_{\text{delete}}$.

\item[(iv)] \para{$\mathcal{L}_{\text{drag}}$.}\
Given source point $A$ and target point $B$, we progressively copy-paste features around $A$ along a linear path towards $B$.
Hence, unlike the other operators, drag involves multiple iterations.
Inspired by~\cite{pan2023drag, shi2023dragdiffusion, mou2023dragondiffusion}, we consider two steps in each iteration to define $\Gamma$ and $V$,~\ie, motion supervision and point tracking.
In the motion supervision step, at the $k$-th iteration, we denote $P_k$ as the source point  (initially $P_0=A$) and $\Gamma(P_k, r_1)$ as the local neighborhood around $P_k$, where $r_1$ is a radius parameter.
Denoting $u$ as the unit vector from $P_k$ to $B$, we translate the local neighborhood around $P_k$ by vector $u$ to locate the target region $\Gamma=\{p+u|p\in \Gamma(P_k, r_1) \}$ and set $V=F_{k}[\Gamma(P_k, r_1)]$ as the target values.
With $\Gamma$ and $V$,
we can then set up objective $\mathcal{L}_{\text{drag}}$ for performing the 
co-optimization at the $k$-th iteration.
In the point tracking step, after the $k$-th iteration, we need to update position $P_k$ to $P_{k+1}$. Importantly, if the source point is not accurately tracked, the next motion supervision step will be supervised by a wrong position, leading to undesired results.
Here, we search for $P_{k+1}$ within a radius parameter $r_2$ around $P_k$,~\ie, $\Gamma(s_k, r_2)$, 
such that its features are most similar to those around source point $A$ in the original neural volume $F_0$:
\end{itemize}
\begin{equation}
\label{eq:drag_track}
   P_{k+1} = \underset{q \in \Gamma(P_k, r_2)}{\text{argmin}} | F_{k+1}[q] - F_0[A] |_1.
\end{equation}

\section{Results and Experiments}
\subsection{Dataset and Implementation Details}
\label{subsec:imp_detail}

We perform evaluations on the following datasets: ShapeNet~\cite{chang2015shapenet}, 
ABC~\cite{koch2019abc}, and SMPL~\cite{SMPL:2015}.
The diffusion U-Net has 16 layers in the output blocks and we take the output of the fourth-last layer 
as the neural feature volume $F$; other choices are ablated in Section~\ref{subsec:ablation}.
For the CNS optimization, we use the Adam optimizer with a learning rate of $3\times 10^{-2}$.
Our diffusion model's inference process spans $T$ = 1000 time steps, and we extract the neural feature volume $F$ specifically at the $t=200$ time step.
The diffusion inference takes around 1 minute, whereas the CNS optimization takes less than 10 seconds for each operator.
For the drag operator, we set $r_1=1$ and $r_2=2$, and its optimization termination condition is met when the source point reaches the target point. For the other three operators, the optimization terminates, once the loss reduces below one-third of its initial value.
For all four operators, the maximum number of optimization steps is capped at 300. 
\emph{We will release code and data after the publication.}

\begin{table}[t]
    \begin{minipage}[t]{.49\textwidth}
        \scriptsize
    	\centering
    		\caption{
      Quantitative comparisons between our method and other state-of-the-art methods. We can see that the edited shapes generated by our method have the best quality for all the metrics: lowest Frechet Inception Distance (FID), lowest Kernel Inception Distance (KID), highest Quality Score (QS), and highest Matching Score (MS). Since DeepMetaHandle~\cite{liu2021deepmetahandles} does not offer pre-trained models for the airplane class, a comparison with this method is not applicable to the airplane class.}
      \vspace{-3mm}
    	\resizebox{0.99\linewidth}{!}{
    \begin{tabular}{C{1.6cm}|@{\hspace*{0.0mm}}C{0.5cm}@{\hspace*{0.0mm}}@{\hspace*{0.0mm}}C{0.6cm}@{\hspace*{0.0mm}}C{0.5cm}@{\hspace*{0.0mm}}C{0.5cm}|@{\hspace*{0.0mm}}C{0.5cm}@{\hspace*{0.0mm}}@{\hspace*{0.0mm}}C{0.6cm}@{\hspace*{0.0mm}}@{\hspace*{0.0mm}}C{0.5cm}@{\hspace*{0.0mm}}@{\hspace*{0.0mm}}C{0.5cm}@{\hspace*{0.0mm}}@{\hspace*{0.0mm}}C{0.5cm}@{\hspace*{0.0mm}}}
    \cline{2-9}
    \multicolumn{1}{c|}{} & \multicolumn{4}{c|}{Chair}                                  & \multicolumn{4}{c}{Airplane}                                \\ \hline
    Method                & FID $\downarrow$ & KID $\downarrow$ & QS $\uparrow$ & MS $\uparrow$ & FID $\downarrow$ & KID $\downarrow$ & QS $\uparrow$ & MS $\uparrow$ \\ \hline
    DeepMetaHandle       & 118.2                  & 0.028                  &  3.40  & 1.78   &  \multicolumn{1}{c}{-}    & -    &  -  & -   \\
    SLIDE                 & 100.4                  & 0.012                  &   3.65 &  1.86  & 127.6                 & 0.043                  &  3.13  &  1.71  \\
    DualSDF               & 122.9                  &  0.018                 &  3.02  &  2.15  &  152.1              & 0.063                  &  2.20  &  1.56  \\
    SPAGHETTI             &   145.9        &         0.036          &  2.98  &  2.36  & 179.1    &  0.088                &  1.62  & 2.59   \\ \hline
    CNS-Edit (ours)                  & \textbf{88.7}                   & \textbf{0.006}                  &  \textbf{4.50}   &  \textbf{4.59}  & \textbf{106.9}                & \textbf{0.034}                  &  \textbf{4.11}  & \textbf{4.06}  \
    \end{tabular}
            }
    	\label{tab:drag_compare}
    	\vspace{-2mm}
    \end{minipage}
    \hfill
\end{table}

\subsection{Experiment Settings}
We developed four operators: drag, delete, resize, and copy. However, existing methods largely overlooked the last three operators. Hence, our primary focus is on evaluating and comparing the performance of our drag operator with others. 
We compare our method against four shape editing methods. Among these methods, DualSDF~\cite{hao2020dualsdf}, SPAGHETTI~\cite{hertz2022spaghetti}, and SLIDE~\cite{lyu2023controllable} couple the 3D shapes with corresponding coarse geometric primitives, so shape editings are performed by moving these primitives.
Besides, we compare our method with DeepMetaHandle~\cite{liu2021deepmetahandles}, which utilizes a deformation network to predict the vertices offset to edit the shapes.
To facilitate the comparison, we follow~\cite{hu2023clipxplore, liu2022iss} to build a dataset comprising 50 editing cases, which include 25 chairs and 25 airplanes from the ShapeNet dataset.
In particular, we ask one participant to decide the pair of source and target points for each shape. Next, we evaluate our method and other baselines on this dataset.

\subsection{Quantitative Comparison}

\paragraph{Evaluation metrics} 
First, we employed the Frechet Inception Distance (FID) and Kernel Inception Score (KID) to evaluate the visual quality of the shapes produced by different methods.
Evaluating the ability of the method to produce shapes that match the user operation poses a challenge. To this end, we conducted a user study to assess the fulfillment of the user operation
thoroughly.
Following~\cite{hu2023clipxplore, liu2022iss}, we invited 10 participants to evaluate the edited shapes, focusing on two key aspects: the Quality Score (QS), which assesses the visual appeal of the shapes, and the Matching Score (MS), which determines how well they align with the user operation. For each edited shape, we asked each participant to give ratings on QS and MS, ranging from 1 (worst) to 5 (best).
Table~\ref{tab:drag_compare} demonstrates that our method outperforms existing state-of-the-art methods. The MS score in Table~\ref{tab:drag_compare} shows that our method is more effective in producing shapes that match the user operation. Additionally, our method generates edited shapes of higher quality and fidelity, as confirmed by the FID, KID, and QS metrics.

\subsection{Qualitative Comparison}
\begin{figure*}[h]
\vspace*{-1mm}
  \centerline{\includegraphics[width=0.99\linewidth]{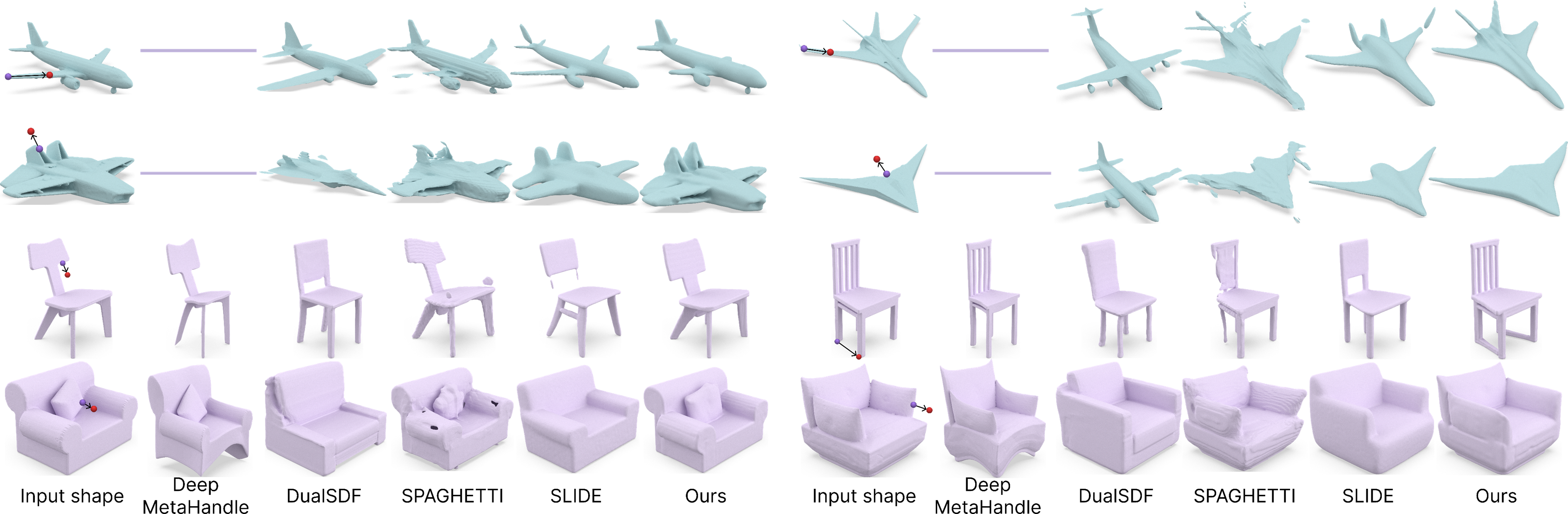}}
\vspace*{-3.5mm}
\caption{
Visual comparisons between our method and state-of-the-art methods:
DeepMetaHandle~\cite{liu2021deepmetahandles}, DualSDF~\cite{hao2020dualsdf}, SPAGHETTI~\cite{hertz2022spaghetti}, and SLIDE~\cite{lyu2023controllable}. 
Our method enables topology modification while the deformation-based method~\cite{liu2021deepmetahandles} struggles with it. Also, our method maintains shape semantics during the editing. For example, as the last example on the right shows, when one armrest of the sofa is dragged, the other armrest would adjust automatically. Our method can produce edited shapes that better match the editing operation with less visual artifacts. 
}
\label{fig:drag_compare}
\end{figure*}

Figure~\ref{fig:drag_compare} shows the visual comparison between our method and other baselines. 
We can achieve shape editing with semantics by leveraging the coupled neural shape representation, as demonstrated by the airplane example shown in the first row of Figure~\ref{fig:drag_compare}.
When shortening the left wing of the airplanes (top two examples), the right wing would shorten automatically, reflecting that our method incorporates shape semantics,~\eg, symmetry, into the editing process.
Meanwhile, other baseline methods often result in implausible shapes or fail to accurately match the specified edit,

Additionally, our method allows for more precise shape editing 
compared with other methods, as showcased in the sofa example in the fourth row on the left.
Our approach enables dragging the corner of the sofa's pillow to reach the sofa's seat, maintaining the pillow's integrity throughout the editing. In contrast, other methods either lose the pillow during the edit or yield a broken pillow.
Moreover, our method is capable of introducing topology modifications during the editing, which is not achievable by deformation-based method~\cite{liu2021deepmetahandles}. This capability is demonstrated on the right side of the third row in Figure~\ref{fig:drag_compare}, where we connect the leg of the chair and modify the topology of the input shape.
For more visual comparisons, please refer to the supplementary material.

\subsection{More Visual Results for Different Operators}
\begin{figure*}[h]
\vspace*{-1mm}
  \centerline{\includegraphics[width=0.99\linewidth]{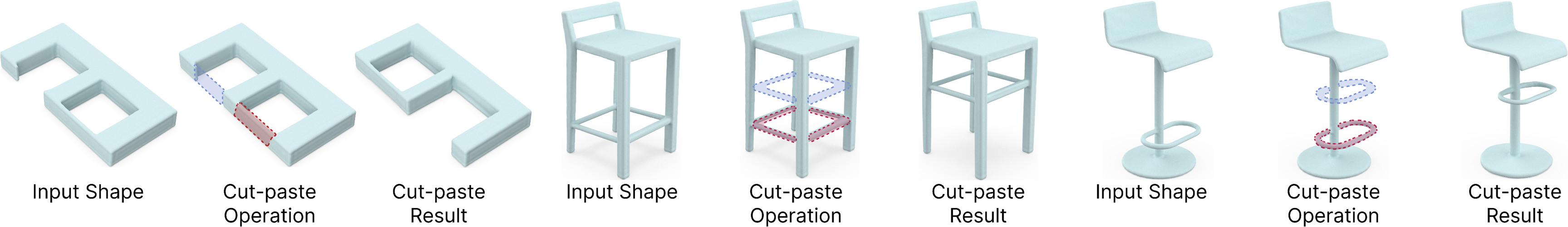}}
\vspace*{-3.5mm}
\caption{
Visual results of the cut-paste operator. Our method enables one to choose a part (in red) of a shape, then cut-and-paste it to a target region (in blue).
}
\label{fig:gallery_cut}
\end{figure*}

\begin{figure*}[h]
\vspace*{-1mm}
  \centerline{\includegraphics[width=0.99\linewidth]{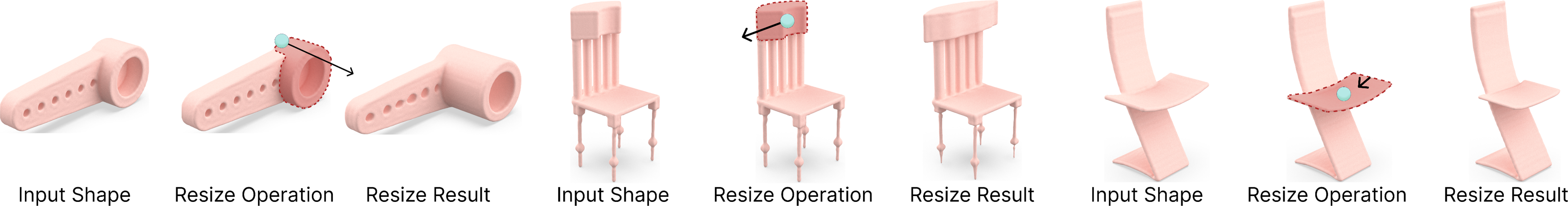}}
\vspace*{-3.5mm}
\caption{
Visual results of the resize operator. Our method enables one to select a region (in red) and resize it along a certain axis (black arrow) about a chosen anchor point (blue dot). The edited results are produced by matching the resize operations.
}
\label{fig:gallery_resize}
\end{figure*}

\begin{figure*}[h]
\vspace*{-1mm}
  \centerline{\includegraphics[width=0.99\linewidth]{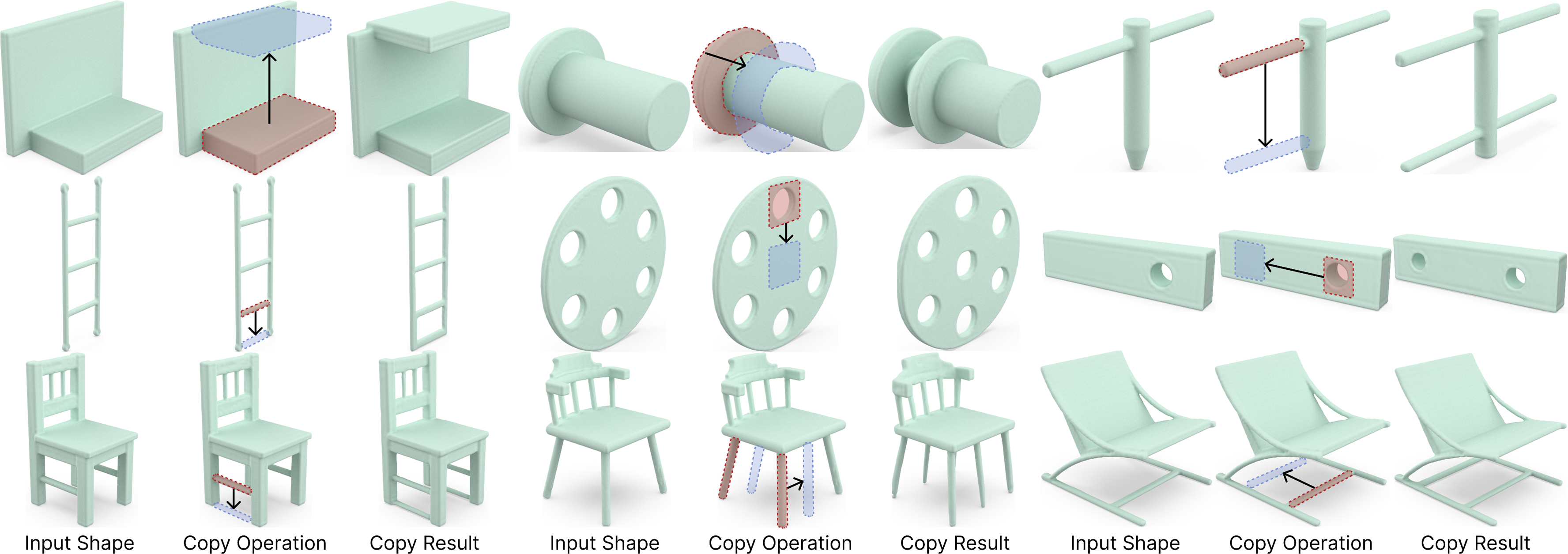}}
\vspace*{-3.5mm}
\caption{
Visual results for the copy operator. Our method enables one to choose a source region (in red) in a shape and then paste the local geometries to a target region (in blue).
}
\label{fig:gallery_copy}
\end{figure*}

\begin{figure*}[h]
\vspace*{-1mm}
  \centerline{\includegraphics[width=0.99\linewidth]{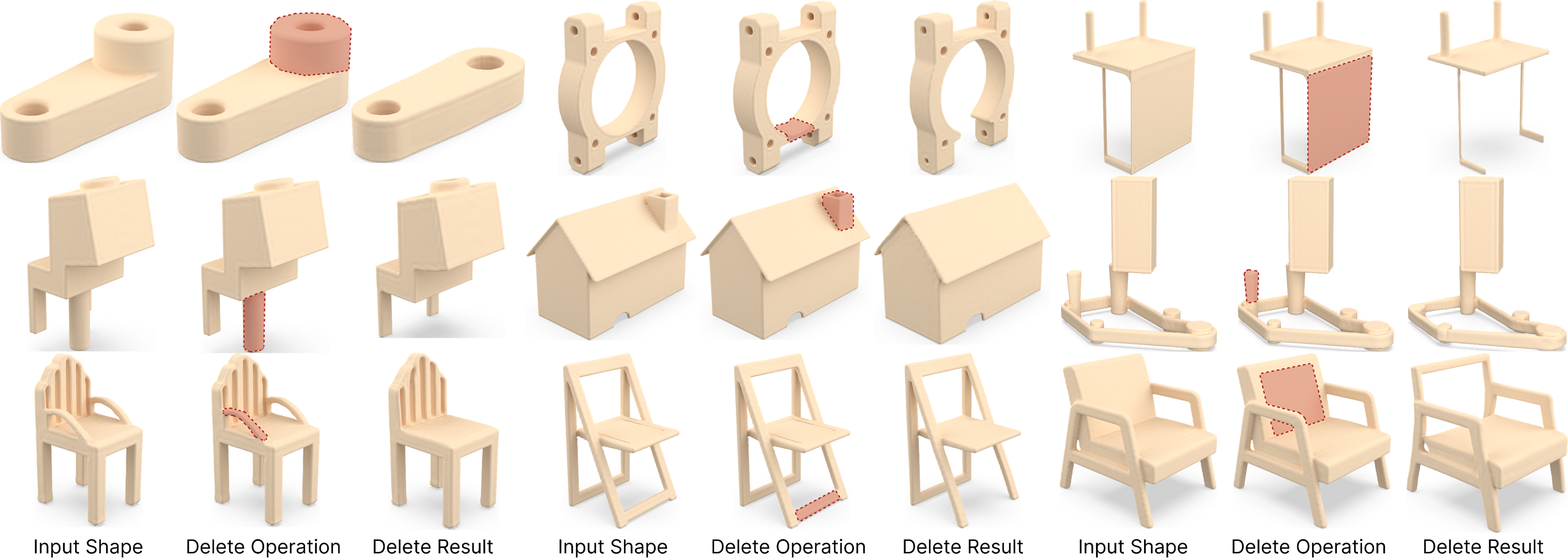}}
\vspace*{-3.5mm}
\caption{
Visual results for the delete operator. Our method allows one to choose a
part (in red) of a shape and remove it.
}
\label{fig:gallery_delete}
\end{figure*}

\begin{figure*}[h]
  \centerline{\includegraphics[width=0.99\linewidth]{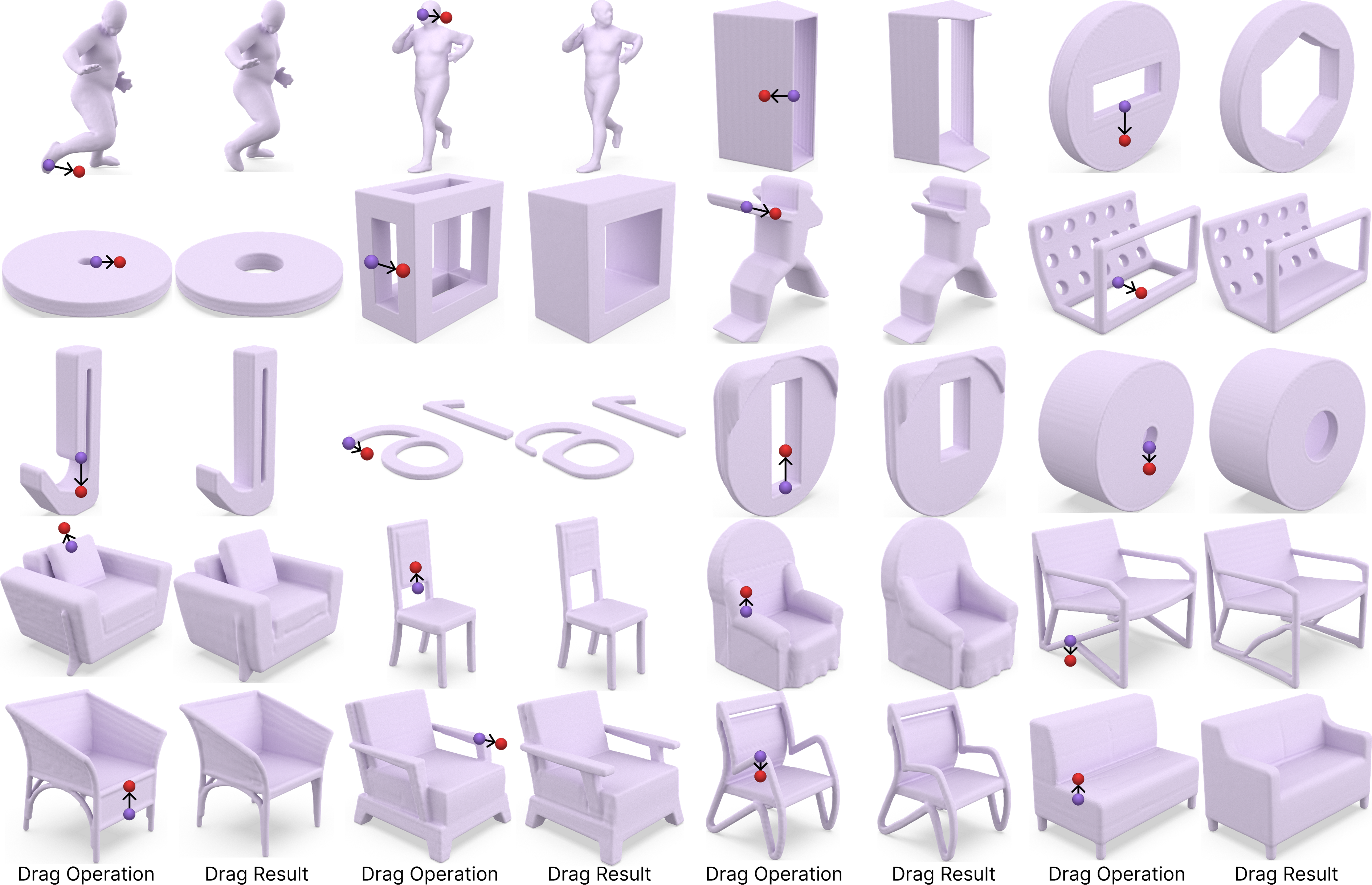}}
\vspace*{-3.5mm}
\caption{
Visual results for the drag operator. By providing a pair of source point (in purple) and target point (in red), 
our method is capable of modifying a diverse range of shapes to produce high-fidelity results that match the drag operations.
}
\label{fig:gallery_drag}
\end{figure*}

We provide additional visual results of our proposed operators,
including resize, copy, delete, and drag, as shown in Figures~\ref{fig:gallery_resize},~\ref{fig:gallery_copy},~\ref{fig:gallery_delete}, and ~\ref{fig:gallery_drag}.
Furthermore, we present additional examples of the cut-paste operator, which combines the copy and delete operators, as depicted in Figure~\ref{fig:gallery_cut}.
Beyond the ShapeNet dataset~\cite{chang2015shapenet}, we apply our operators to a CAD dataset~\cite{koch2019abc}, which involves shapes with diverse topologies and without a canonical orientation as ShapeNet. Notably, our operators function effectively with CAD models, demonstrating the robustness of CNS-Edit. 

Though our editing operator typically preserve the unedited regions, it sometimes may slightly alter regions with fine geometric details. To protect these regions from unwanted changes, we adopt the region-aware constraint method in~\cite{hu2023neural} to effectively preserve the integrity of specified regions during the editing. For more information on the adoptation of this method and further visual results, please refer to the supplemental materials.

\subsection{Ablation Study}
\label{subsec:ablation}
\begin{figure}[!t]
	\centering
	\includegraphics[width=0.99\columnwidth]{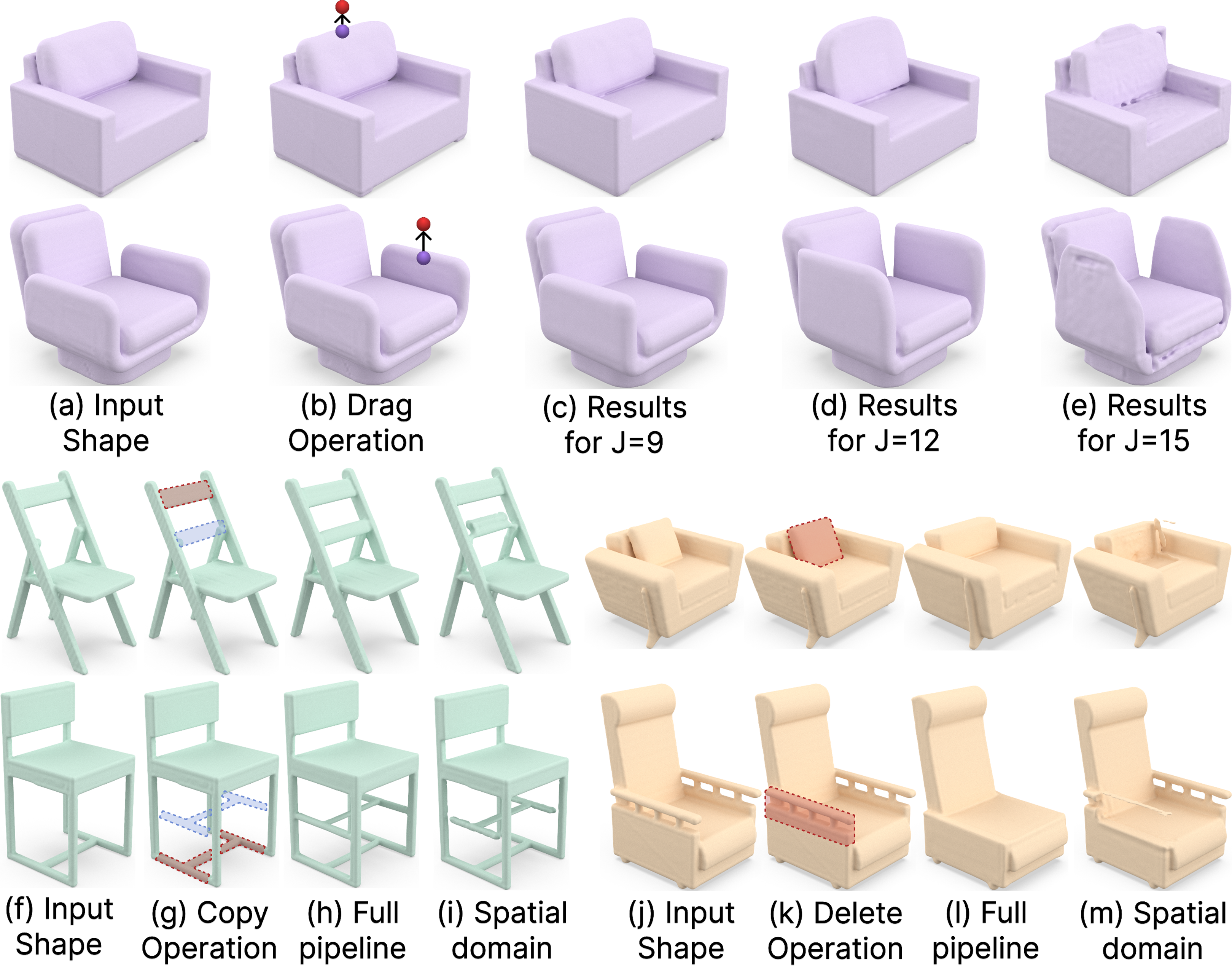}
        \vspace{-2mm}
	\caption{
 Visual results from the ablation study.
 Using features closer to the output when constructing neural volume $F$ introduces artifacts in the edited shapes, as seen in (e). However, features from too deep layers lack spatial context, resulting in less effective editing, as evident in (c). Further, applying our operators directly in the spatial domain leads to a loss of shape semantics during editing, compare (l) \& (m), and also causes artifacts in the edited shapes, noticeable in (i) and (m) vs. (h) and (l), correspondingly.}
	\label{fig:demo_abl}
\end{figure}

We first ablate the effects of utilizing features from different layers of the U-Net (16 layers in total) to create the neural volume $F$.
In our standard setting, we use the feature volume output from the 12-th layer (denoted as $J=12$) to construct $F$.
To conduct the ablation study, we experiment with features from shallower layers ($J=15$) and deeper layers ($J=9$) for constructing $F$. 
Features extracted from shallow layers ($J=15$) provide rich spatial context, yet they contain limited shape semantics. This limitation is evident in Figure~\ref{fig:demo_abl} (e), where utilizing features from these shallow layers enables results to match the editing operation. However, due to the absence of shape semantics, the quality of the edited shapes is compromised, often resulting in 
obvious
artifacts.
Features from deeper layers ($J=9$) can provide more detailed and abstract representations, but they clearly lack 
spatial context. The loss of spatial context will affect the controllability during the editing. 
As Figure~\ref{fig:demo_abl} (c) shows, using features from deeper layers ($J=9$) results in edited shapes that strongly resemble the input shape, without much changes. This similarity indicates that the features from the deeper layers, being more abstract, lack sufficient spatial context. Consequently, this leads to a reduced level of controllability in the editing process.

Next, we conduct another ablation by directly applying our operators in the spatial domain,~\ie, wavelet volume. 
Specifically, instead of applying our CNS co-optimization procedure, we directly assign the corresponding target values to the target region in the wavelet volume.
For the copy operator, this entails substituting the values in the target pasted region with those from the corresponding source copy region in the wavelet volume. 
In the case of the delete operator, we substitute the values in the selected region with those from an empty region.
As Figures~\ref{fig:demo_abl} (i) and (m) show, directly applying the operator within the spatial domain can lead to artifacts in the edited shapes. 
Furthermore, this direct application of the operators may also lack an understanding of shape semantics during editing, as demonstrated in Figure~\ref{fig:demo_abl} (m).
It is evident that the application of the proposed operators in the neural volume domain yields edited results with high fidelity and the editing process can also maintain the shape semantics; see Figures~\ref{fig:demo_abl} (h) and (l).

\section{Limitations and Future Work}
Although our method allows for high-fidelity shape editing with various operators provided, there are still some limitations.
First, our method builds upon the pre-trained space latent space introduced by~\cite{hu2023neural} for constructing the coupled neural shape (CNS) representation. However,~\cite{hu2023neural} is category-specific, requiring the training of separate models for different categories of 3D shapes. This constraint inhibits its ability to reconstruct and subsequently edit arbitrary 3D shapes. 
Recently, some work~\cite{ren2023xcube} aims to reconstruct 3D shapes by training on large scale 3D data~\cite{deitke2023objaverse, deitke2023objaversexl}. So, one future direction is to edit arbitrary shapes by integrating these models with our approach.

Moreover, though our current approach for shape editing is effective, it still faces a bottleneck in terms of processing time. The coupled neural shape optimization, operates efficiently, completing optimization in under 10 seconds per operator.
However, the diffusion-based shape reconstruction network~\cite{hu2023neural} requires a minute to generate edited shapes from the updated latent code. Therefore, one possible direction is to integrate a more efficient shape reconstruction framework with our method. 

Lastly, while our method currently supports four atomic operators, it is capable of creating new ones,~\eg, cut-and-paste, simply by combing the atomic operators. That said, there are still editing operations that we have not explored. Table~\ref{tab:limitation} summarizes the editing operators supported by SPAGHETTI and CNS-Edit, respectively. Note that our current method does not explore operators such as rotation. Also, though our current CNS representation shows strong capacity in single-shape editing, we have not studied blending different shapes,~\eg, shape mixing and part interpolation. We will leave shape blending and rotation for future work.

\begin{table}[!t]
    \begin{minipage}[t]{.49\textwidth}
        \scriptsize
    	\centering
    		\caption{
      Single-shape editing operators supported by ours and SPAGHETTI.}
      \vspace{-3mm}
    	\resizebox{0.99\linewidth}{!}{
    \begin{tabular}{C{1.0cm}|@{\hspace*{0.0mm}}C{0.5cm}@{\hspace*{0.0mm}}@{\hspace*{0.0mm}}C{0.8cm}@{\hspace*{0.0mm}}C{0.8cm}@{\hspace*{0.0mm}}C{0.8cm}@{\hspace*{0.0mm}}C{0.8cm}@{\hspace*{0.0mm}}@{\hspace*{0.0mm}}C{1.0cm}@{\hspace*{0.0mm}}@{\hspace*{0.0mm}}C{2.0cm}@{\hspace*{0.0mm}}}
    \cline{2-7}
                    & Drag & Copy & Resize & Delete & Rotate & Cut-paste  \\ \hline
    SPAGHETTI             &   $\checkmark$       &         $\times$          &  $\times$  &  $\times$  & $\checkmark$    &  $\times$                   \\ 
    Ours                  & $\checkmark$                   & $\checkmark$                  &  $\checkmark$   &  $\checkmark$  & $\times$                & $\checkmark$                   \
    \end{tabular}
            }
    	\label{tab:limitation}
    \end{minipage}
    \hfill
\end{table}

\vspace*{4mm}
\section{Conclusion}
In this work, we propose a new coupled neural shape (CNS) representation, consisting of a latent code $z$ and corresponding neural volume $F$.
Based on this CNS representation, we propose a family of operators,~\ie, copy, resize, delete, and drag operators, that can be applied to the CNS representation for shape editing. 
Given the specific operator, we aim to convert it into a corresponding objective function. Then we can apply the objective to our CNS representation to co-optimize the latent code $z$ and neural volume $F$ to fulfill the given editing operator. Next, we can decode the modified CNS representation to obtain the edited shape.
Our experiments demonstrate that our method can produce edited shapes with high fidelity and match the desired shape changes, surpassing the existing works.

\bibliographystyle{ACM-Reference-Format}
\bibliography{bibliography}

\clearpage

\end{document}